\documentclass[conference]{IEEEtran}
\IEEEoverridecommandlockouts
\usepackage{cite}
\usepackage{amsmath,amssymb,amsfonts}
\usepackage{graphicx}
\usepackage{textcomp}
\usepackage{xcolor}
\usepackage{algorithm}
\usepackage{algpseudocode}
\usepackage{pgfplots}
\usepackage[T1]{fontenc}
\pgfplotsset{width=0.45\textwidth,compat=1.9}
\def\BibTeX{{\rm B\kern-.05em{\sc i\kern-.025em b}\kern-.08em
    T\kern-.1667em\lower.7ex\hbox{E}\kern-.125emX}}

\author{\IEEEauthorblockN{Marcus Rüb}
\IEEEauthorblockA{\textit{Hahn-Schickard}\\
Villingen-Schwenningen, Germany \\
Marcus.Rueb@Hahn-Schickard.de}
\and
\IEEEauthorblockN{Daniel Konegen}
\IEEEauthorblockA{\textit{Hahn-Schickard}\\
Villingen-Schwenningen, Germany \\
Daniel.Konegen@Hahn-Schickard.de}
\and
\IEEEauthorblockN{Patrick Selle}
\IEEEauthorblockA{\textit{Hahn-Schickard}\\
Villingen-Schwenningen, Germany \\
Patrick.Selle@Hahn-Schickard.de}
\and
\IEEEauthorblockN{Axel Sikora}
\IEEEauthorblockA{\textit{Offenburg University}\\
Offenburg, Germany \\
\textit{Hahn-Schickard}\\
Villingen-Schwenningen, Germany \\
Axel.Sikora@hs-offenburg.de}
\and
\IEEEauthorblockN{Daniel Mueller-Gritschneder}
\IEEEauthorblockA{\textit{Embedded Computing Systems} \\
\textit{Faculty of Informatics, TU Wien}\\
Vienna, Austria \\
daniel.mueller-gritschneder@tuwien.ac.at}

}

\begin{document}

\title{DRIP: DRop unImportant data Points - Enhancing Machine Learning Efficiency with Grad-CAM-Based Streaming Data Prioritization for On-Device Training}

\maketitle

\begin{abstract}
Selecting data points for model training is critical in machine learning. Effective selection methods can reduce the labeling effort, optimize on-device training for embedded systems with limited data storage, and enhance the model performance. This paper introduces a novel algorithm that uses Grad-CAM to make online decisions about retaining or discarding data points. Optimized for embedded devices, the algorithm computes a unique DRIP Score to quantify the importance of each data point. This enables dynamic decision-making on whether a data point should be stored for potential retraining or discarded without compromising model performance. Experimental evaluations on four benchmark datasets demonstrate that our approach can match or even surpass the accuracy of models trained on the entire dataset, while achieving storage savings of up to 39\%. To our knowledge, this is the first algorithm to make online decisions about data point retention without requiring access to the entire dataset.

\end{abstract}


\section{Introduction}

In the rapidly evolving domain of machine learning, the quantity of available data have reached unprecedented levels. While large datasets have traditionally been the bedrock of robust machine learning models, the sheer magnitude of data now available poses both opportunities and challenges. One of the primary challenges is efficient data management, especially but not only in scenarios with constrained computational and storage resources \cite{10650122}. The indiscriminate data accumulation can lead to increased storage costs, longer training times, and potential overfitting due to redundant or wrong-labeled data. Moreover, in many real-world applications, especially in edge computing and embedded systems \cite{Rub.2022}, the storage and computational resources are limited. Yet, new approaches do not only execute model inference on the devices but move towards so-called on-device training, which requires to store data points for model retraining. In such applications arises a need for a systematic approach to discern the utility of each collected data point and decide its retention or discard in streaming is needed. Other applications for data selection methods target to reduce labeling effort or improving model performance \cite{rueb2024}. 
\\

The benefits of data selection approaches are manifold. By selectively retaining data, storage requirements can be significantly reduced, making it feasible for on-device storage and processing. This selective retention also translates to computational efficiency, as less data leads to faster retraining cycles and enables timely model updates even on resource-constrained devices. Additionally, by focusing on the most informative data points, the effort, and cost associated with labeling can be significantly minimized, optimizing the overall data preparation process. This strategy also enhances data economy by transmitting only valuable data, reducing data communication overhead, leading to energy savings and prolonged device lifetimes. Furthermore, concentrating on crucial data points allows the model to potentially learn more salient features, thereby improving its overall performance \cite{Wu.02.03.2022}.
\\

To address these challenges, we propose leveraging insights from the model’s internal decision-making process. Specifically, we use Gradient-weighted Class Activation Mapping (Grad-CAM), a technique originally developed to visualize model predictions, as the foundation for quantifying and prioritizing data points during streaming.
This raises the central research question of our study: How can selective, streaming data storage using a Grad-CAM-based scoring system improve model accuracy and reduce memory requirements for machine learning on devices in resource-constrained environments, while reducing labeling overhead without compromising model performance?
\\

This paper delves into this challenge and presents a novel algorithm that employs Grad-CAM to make informed decisions about data retention \cite{Jahmunah.2022}. The motivation for this research stems from the increasing need for efficient data management in embedded machine learning (tinyML) \cite{Plancher}. Traditional data retention methods are inadequate for embedded applications, necessitating a novel approach to ensure both storage efficiency and model accuracy. Our proposed method calculates a distinct metric, termed the DRIP Score (DRIPS), which quantifies the relevance of each data point in the context of model training and performance. Using this metric, the algorithm can dynamically assess the significance of a streaming data point and make decisions about its retention.
\\

The distinct novelty of our approach doesn't only include its online decision-making capability but also in its innovative application of the Grad-CAM technique. While Grad-CAM is a well-established method for visualizing model decisions, its application in the domain of data retention was not yet explored. Furthermore, our method introduces a novel computation derived from the Grad-CAM outputs, offering a fresh perspective on how these visualizations can be quantified and utilized for practical decision-making in data selection, especially for on-device training scenarios\cite{Sharma.2021}.
\\

Our experimental evaluations on four benchmark datasets, MNIST, CIFAR-10, Speech Commands, and Plant Disease, demonstrate that our approach can match or even surpass the accuracy of models trained on the entire dataset, all while achieving storage savings of up to 39\%.

\section{Related Work}
Data valuation intersects with active learning and coreset selection to enhance neural network efficiency. This section explores state-of-the-art (SOTA) techniques in these areas, highlighting their contributions to machine learning.

\subsection{Relevant Works in Data Valuation}

Data valuation techniques aim to quantify the importance of individual data points for model performance, enabling more efficient training and storage. Strumbelj and Kononenko \cite{Strumbelj.2014} brought forth Data Shapley as a method to quantify the importance of individual data points within a dataset. By evaluating each point's contribution to the model's overall performance, Data Shapley offers guidance on data selection for both training and deployment. Despite its potential, Data Shapley is computationally intensive, particularly for large datasets, and operates offline, restricting streaming data importance assessments \cite{Jia.27.02.2019}. Such challenges necessitate continued research to harness its full potential in machine learning tasks.

Data Valuation in Machine Learning: Ingredients, Strategies, and Open Challenges by Sim et al. \cite{Sim.2022} provides a comprehensive survey on data valuation in machine learning, elucidating its ingredients and properties. While the authors present an encompassing view of the topic, they do not introduce a novel algorithm. In contrast, our approach pioneers in leveraging Grad-CAM for online data valuation.

The paper from Wang and Jia \cite{Wang.31.05.2022} emphasizes the robustness of data valuation, advocating for the Banzhaf value from cooperative game theory. The distinct direction of our work lies in the incorporation of Grad-CAM for online decision-making, providing a fresh perspective in data valuation.

The work from Xu et al.~\cite{Xu.uuuuuuuu} focused on the health domain and the pricing from data, this research introduces the Valuation And Pricing mechanism called VAP mechanism for online data valuation. While our method also operates online, it uniquely integrates Grad-CAM for decision-making, thus enriching the data valuation landscape, especially when considering data point significance for the entire dataset.

\subsection{Active Learning}

Active learning seeks to optimize the efficiency of neural network models by selecting the most informative data points for training. Barbulescu \cite{Barbulescu.2023} explored the relative performance of LSTM and GRU architectures, while Guo \cite{Guo.2022} introduced the Recurrent Attention Model (RAM), which integrates reinforcement learning to enhance model performance by focusing on critical regions in input data. While these models offer significant advancements, they rely on complex architectures and do not address the problem of efficient streaming data retention.

Mairittha et al. \cite{nattaya_mairittha__2019} proposed an LSTM-based on-device deep learning inference method that reduces labeling effort by improving data collection quality in human activity recognition systems. While this approach is focused on improving user engagement, our DRIP method operates more broadly, targeting various data types (e.g., images, audio) and offering an automatic, streaming filtering mechanism to discard less informative data points. This makes DRIP more adaptable across diverse datasets beyond activity recognition.

Tharwat and Schenck \cite{alaa_tharwat__2024} tackled missing data in IoT devices using a query selection strategy that accounts for imputation uncertainty. Their active learning method selects representative data points, improving classification even with incomplete data. However, this approach is predominantly offline. In contrast, our DRIP system assesses data importance dynamically in real time, optimizing data retention for embedded systems without requiring prior imputation or handling missing data.

\subsection{Coreset Selection}

Coreset selection focuses on reducing the size of training datasets by identifying the most informative samples while maintaining model performance. Yoon et al. \cite{Yoon.2021} presented an online coreset selection method validated across multiple datasets, improving continual learning performance. Similarly, Guo et al. \cite{Guo.2022} proposed adaptive second-order coresets, which consider data points and their curvature to reduce training data size. While these approaches offer efficient data reduction, they do not provide the on-device, dynamic decision-making capabilities of our DRIP method.

Venkataramani et al. \cite{swagath_venkataramani__2016} explored hardware-based methods like machine learning accelerators and approximate computing to improve computational efficiency in IoT devices. Although these methods are essential for enhancing the hardware capabilities of resource-constrained systems, our approach focuses on software-level efficiency, selectively retaining data points to reduce storage and computational needs. This makes DRIP a complementary solution to hardware improvements.

Coreset selection methods like those by Balles et al. \cite{Balles.2021}, which use gradient matching for selecting subsets of data, and Ju et al. \cite{Ju.2022}, which utilize contrastive learning for unsupervised coreset selection, offer advancements in dataset reduction. However, these methods are designed for offline data management, whereas our DRIP algorithm operates with streaming data, making dynamic retention decisions based on Grad-CAM heatmap analysis.

Moore et al. \cite{Moore.2023} raised concerns about dataset balancing, arguing that it may degrade performance on unseen datasets. Unlike these static methods, our approach continuously evaluates the importance of each data point during training, preventing overfitting and ensuring data retention is context-aware and flexible.

Hong et al. \cite{Hong.2024} introduced the Evolution-aware Variance (EVA) coreset selection for medical image classification, which achieved high compression rates with minimal accuracy loss. While EVA focuses on offline optimization, our method evaluates streaming data importance on-device, providing a more adaptable solution for dynamic environments like embedded systems.

\subsection{Comparison with Our Method}

Our DRIP method is specifically designed for embedded systems and TinyML, where storage and computational resources are limited. Unlike the methods discussed above, which often require significant storage or computational overhead, our approach leverages Grad-CAM to make streaming data retention decisions. This allows us to reduce storage needs while maintaining or improving model performance, making it an ideal solution for resource-constrained environments.

Compared to coreset selection methods that focus on offline data reduction, our approach offers the advantage of making online, context-aware decisions about which data points to retain. This capability enables continuous learning and adaptation in streaming environments, ensuring that only the most informative data points are stored for future training.

Moreover, the versatility of our DRIP method, which can be applied to various data types (images, audio, sensor data), sets it apart from domain-specific methods like the unit selection for text-to-speech synthesis discussed by Karabetsos et al. \cite{Karabetsos.2009}. Our method’s broad applicability makes it suitable for a wide range of machine learning tasks in constrained environments.

\subsection{Grad-CAM: Visualizing Model Decisions}

Introduced by \cite{Selvaraju.op.2017}, Gradient-weighted Class Activation Mapping (Grad-CAM) is a visualization technique used to understand which regions of an input image contribute the most to a neural network's prediction. It provides insights into the decision-making process of convolutional neural networks (CNNs) by producing a heatmap that highlights the influential regions of the input.

The core idea behind Grad-CAM is to compute the gradient of the output score for a target class (before the softmax operation) with respect to the feature maps of a convolutional layer. These gradients serve as weights to produce a weighted combination of the feature maps, resulting in a coarse heatmap of the same size as the feature maps.
Beyond images, \cite{Jahmunah.2022} demonstrated Grad-CAM's applicability to ECG data, underscoring its versatility.

The steps to compute the Grad-CAM heatmap are as follows:

\begin{enumerate}
    \item Let \( y^c \) be the score for class \( c \) (before the softmax). Compute the gradients of \( y^c \) with respect to the feature maps \( A \) of a convolutional layer:
    \begin{equation}
    \frac{\partial y^c}{\partial A^k}
    \end{equation}
    where \( k \) is the index of the feature map.

    \item Perform Global Average Pooling on the gradients to obtain the weights \( \alpha^c_k \) for each feature map:
    \begin{equation}
    \alpha^c_k = \frac{1}{Z} \sum_i \sum_j \frac{\partial y^c}{\partial A^k_{ij}}
    \end{equation}
    where \( Z \) is the number of elements in feature map \( A^k \), and \( i, j \) are spatial indices.

    \item Compute the weighted combination of the feature maps to obtain the raw Grad-CAM heatmap \( L^c_{\text{Grad-CAM}} \):
    \begin{equation}
    L^c_{\text{Grad-CAM}} = \sum_k \alpha^c_k A^k
    \end{equation}

    \item Apply the ReLU activation function to the raw heatmap to obtain the final Grad-CAM heatmap:
    \begin{equation}
    L^c_{\text{Grad-CAM}} = \max(0, L^c_{\text{Grad-CAM}})
    \end{equation}
\end{enumerate}

\subsection{Overview of Evaluated Datasets}
To evaluate the efficacy of our algorithm across diverse domains, we tested it on four distinct datasets, each representing different areas of application:

\begin{enumerate}
\item \textbf{MNIST:} The MNIST database (Modified National Institute of Standards and Technology database) is a renowned collection of handwritten digits. Comprising a training set of 60,000 examples and a test set of 10,000, it is derived from the larger NIST Special Database 3 and Special Database 1. These databases contain monochrome images of handwritten digits from U.S. Census Bureau employees and high school students, respectively. The digits in MNIST have undergone size normalization to fit within a 20x20 pixel box, preserving their aspect ratio, and have been centered in a 28x28 image using the center of mass of the pixels. The normalization process introduces grey levels due to the anti-aliasing technique employed \cite{lecun-mnisthandwrittendigit-2010}.
\item \textbf{CIFAR-10:} The CIFAR-10 dataset, a subset of the Edge Images dataset, consists of 60,000 color images of 32x32 resolution, spread across 10 distinct classes: airplane, automobile, bird, cat, deer, dog, frog, horse, ship, and truck. Each class contains 6,000 images, with a split of 5,000 for training and 1,000 for testing. The dataset's classification criteria ensure that each image distinctly represents its class, is photo-realistic, and contains a single prominent instance of the object \cite{Krizhevsky2009}.

\item \textbf{Plant Disease (PD):} This dataset, an augmented version of the original, comprises approximately 87,000 RGB images of both healthy and diseased crop leaves, categorized into 38 classes. The dataset maintains an 80/20 split for training and validation, preserving the directory structure. Additionally, a separate directory with 33 test images was curated for prediction purposes. In our study, we focused exclusively on the tomato classes within this dataset. This subset includes images of tomato leaves affected by various diseases as well as healthy leaves, providing a comprehensive dataset for tomato disease classification \cite{Ruth.2022}.

\item \textbf{Speech Commands (SC):} This audio dataset contains spoken words, tailored to aid in the training and evaluation of keyword spotting systems. It contains over 100,000 one-second recordings of 35 spoken words, recorded at a sampling rate of 16 kHz. Unlike conventional datasets designed for full-sentence automatic speech recognition, this dataset poses unique challenges and requirements. It provides a methodology for reproducible accuracy metrics and describes the data collection and verification process \cite{Warden.10.04.2018}.
\end{enumerate}

\section{Proposed Algorithm}
\begin{figure*}[!hbt] 
	\centering
	\includegraphics[width=0.7\textwidth]{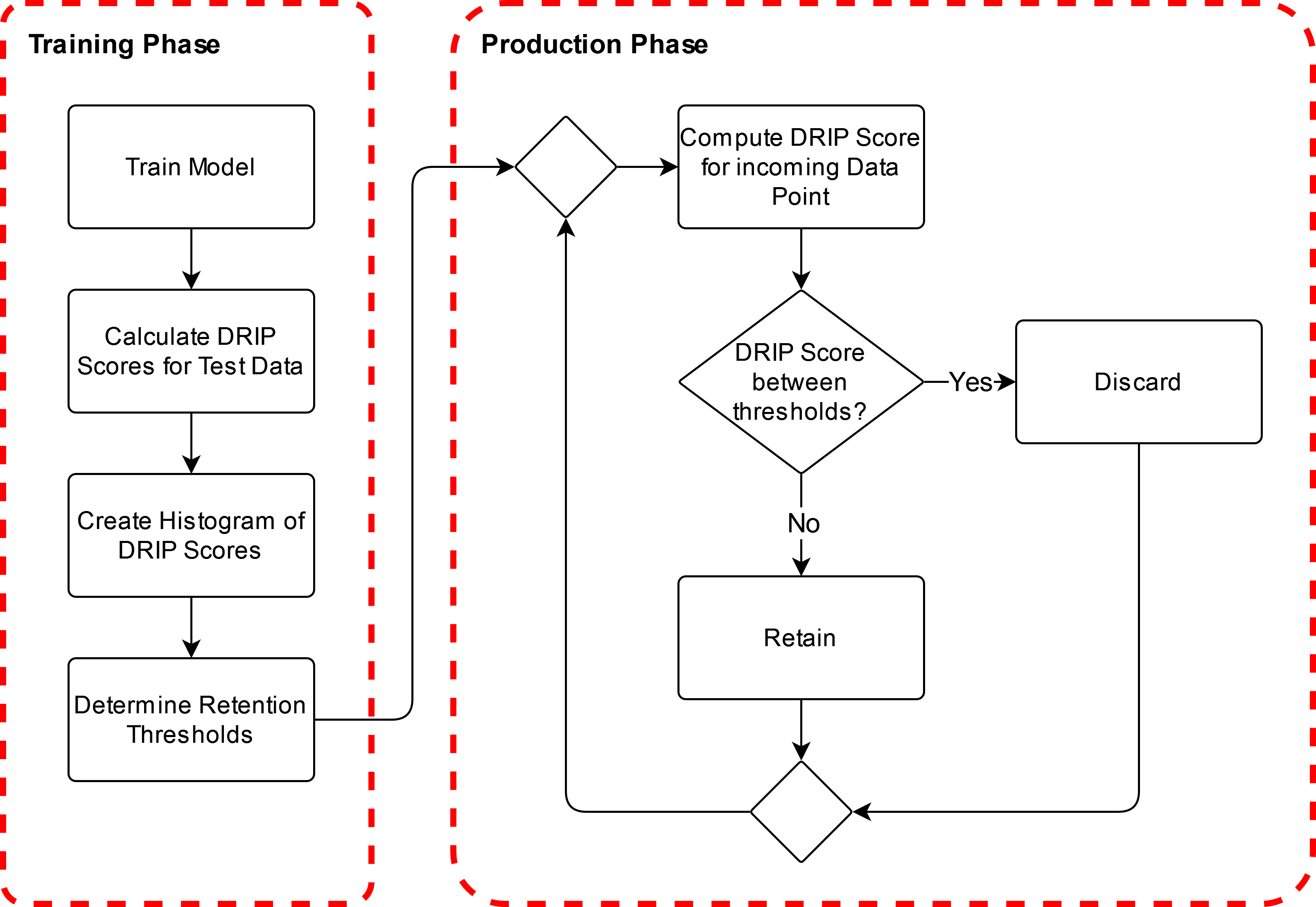} 
\caption{Flowchart illustrating the seven-step process of the proposed DRIP algorithm. The flowchart provides a visual representation of the algorithm's sequential steps, from initial model training to the final decision on data point retention in on-device scenarios.}
\label{fig:algorithm_flow}

\end{figure*}
In the realm of Edge ML, especially in scenarios where data storage and computational resources are constrained, the ability to selectively retain informative data points becomes paramount. The proposed algorithm leverages the Grad-CAM technique, a visualization method designed to highlight regions in an image that a neural network deems important for its predictions. By quantifying the importance of these regions, we can make informed decisions about which data points to retain for potential retraining and which to discard as redundant or routine.

The algorithm operates in two distinct phases: the \textit{Training Phase} and the \textit{Production Phase}. The Training Phase establishes thresholds based on the distribution of DRIP Scores in a test dataset. These thresholds are then utilized in the Production Phase to evaluate incoming data points on-device, deciding their retention based on their computed importance.

\begin{itemize}
    \item \textbf{Training Phase:} This phase involves training the neural network model, computing DRIP Scores for a test dataset, and establishing retention thresholds based on the distribution of heatmap values.
    
    \item \textbf{Production Phase:} In this phase, for each new data point encountered in a production environment, its Grad-CAM heatmap is computed. The data point's retention is then decided based on whether its DRIP Score falls within the thresholds established during the Training Phase.
\end{itemize}

The subsequent subsections provide a detailed breakdown of each phase, elucidating the steps involved and the rationale behind them. Fig.~\ref{fig:algorithm_flow} provides a graphical overview of the sequence of the DRIP algorithm.

\subsection{Training Phase}
\begin{figure}[!hbt] 
	\centering
	\includegraphics[width=0.4\textwidth]{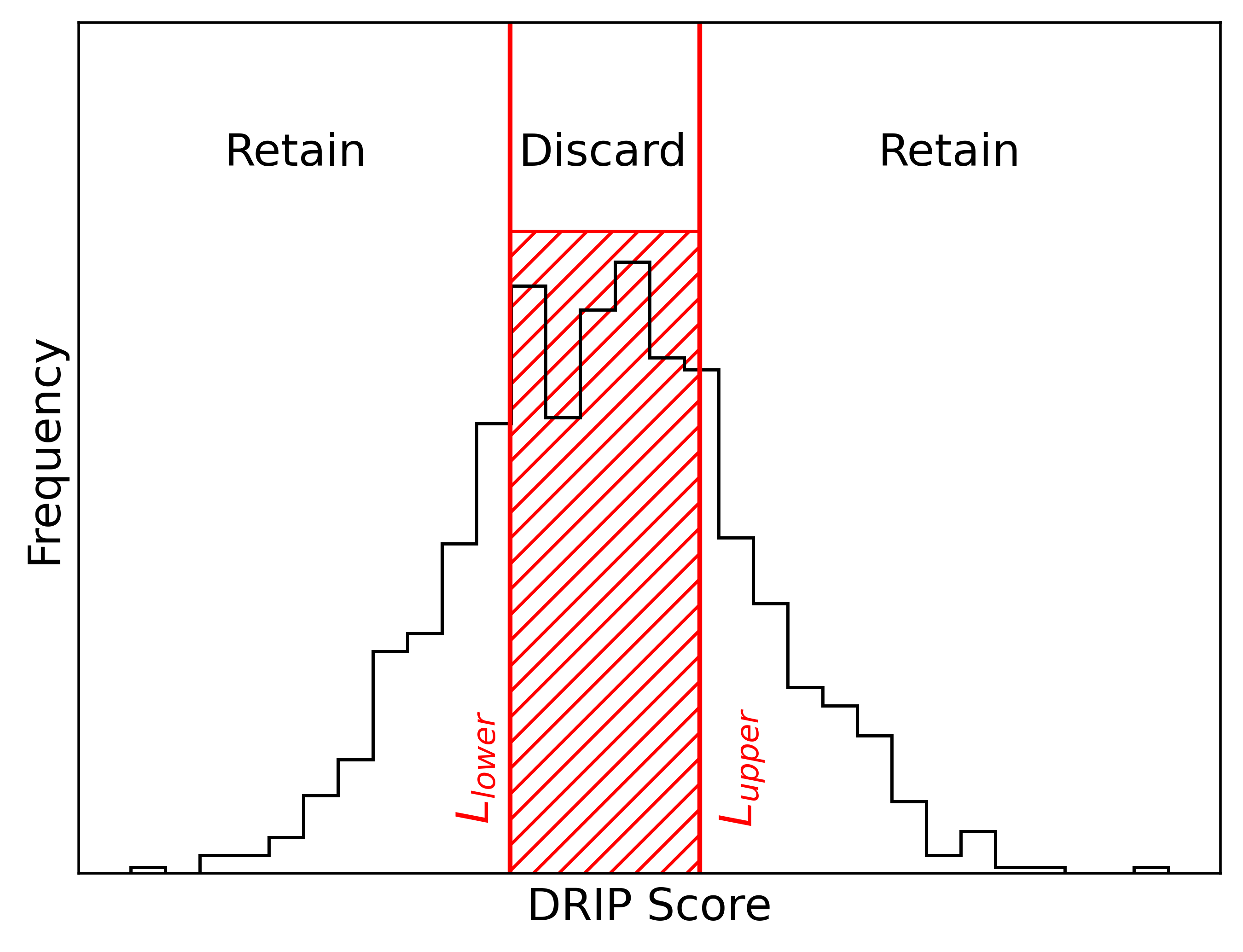} 
    \caption{Determination of retention thresholds from an exemplary DRIP Scores. The peak represents the highest accumulation of DRIP Scores. The calculated lower (\( L_{\text{lower}} \)) and upper (\( L_{\text{upper}} \)) limits encapsulate 25\% of the DRIP Scores, serving as the criteria for our algorithm's data retention decisions.}
    \label{fig:drip_score_determine_retention_thresholds}
    \end{figure}

The Training Phase is crucial for establishing the thresholds that will be used in the Production Phase to determine the importance of a data point. The steps are as follows:

\begin{enumerate}
    \item \textbf{Model Training:} 
    Train a neural network model using the training dataset.

    \item \textbf{Compute DRIP Score (DRIPS):} 
    For each image \( I \) in the test dataset, compute the Grad-CAM heatmap \( H(I) \). For each Grad-CAM heatmap \( H(I) \), calculate the average value of the heatmap. This is done using the formula:
    \begin{equation}
    \text{DRIPS}(I) = \frac{\sum_{i,j} H(I)_{i,j}}{W \times H}
    \end{equation}
    where \( W \) and \( H \) are the width and height of the image \( I \), respectively, and \( i, j \) are pixel indices.
    
    \item \textbf{Histogram Creation:} 
    Construct n histograms (n represents the number of classes in the dataset) using the DRIPS values of all the images in the test dataset. These histograms will show the distribution of DRIPS values across the dataset. 
    
    \item \textbf{Determine Retention Thresholds}: This step involves identifying consistent regions within the DRIPS distributions, assumed to contain less informative data points. The process is as follows:

    \begin{enumerate}
        \item \textbf{Sort DRIPS}: For each class, sort the DRIPS in ascending order.
        
        \item \textbf{Define Discard Percentage Window (DPW)}: Set a window size as a percentage of the total number of scores.
        
        \item \textbf{Slide Window and Calculate Standard Deviation}: Slide this window across the sorted scores. At each position, calculate the standard deviation of the scores within the window.

        \item \textbf{Find Minimum Standard Deviation Window}: Locate the window where the standard deviation is minimal. This window presumably contains the least informative scores.

        \item \textbf{Set \( L_{\text{lower}} \) and \( L_{\text{upper}} \)}: The thresholds are set based on this window's minimum and maximum scores. The lower threshold \( L_{\text{lower}} \) is the minimum score in this window:
        \begin{equation}
        L_{\text{lower}} = \min \left( \text{DRIPS in Selected Window} \right)
        \end{equation}
        The upper threshold \( L_{\text{upper}} \) is the maximum score in the window:
        \begin{equation}
        L_{\text{upper}} = \max \left( \text{DRIPS in Selected Window} \right)
        \end{equation}
        This is shown in Fig.~\ref{fig:drip_score_determine_retention_thresholds}.

        \item \textbf{Thresholds for Each Class}: Repeat steps (a)-(e) for each class to calculate class-specific \( L_{\text{lower}} \) and \( L_{\text{upper}} \).
    \end{enumerate}
\end{enumerate}

The calculated limits \( L_{\text{lower}} \) and \( L_{\text{upper}} \) will serve as thresholds during the Production Phase to decide whether to retain or discard a data point.

\subsection{Production Phase}

Once the thresholds have been established in the Training Phase, the Production Phase uses these to decide the importance of incoming data points. The steps for this phase are:

\begin{enumerate}
    \item \textbf{Compute DRIPS for New Data Point:} 
    For a new data point \( I_{\text{new}} \) processed in production, compute its Grad-CAM heatmap \( H(I_{\text{new}}) \). To ensure the heatmap accurately reflects the areas influencing the model's prediction, the predicted label obtained from the model is used. Subsequently, calculate the DRIPS for this heatmap using the formula: 
    \begin{equation}
    \text{DRIPS}(I_{\text{new}}) = \frac{\sum_{i,j} H(I_{\text{new}})_{i,j}}{W \times H}
    \end{equation}
    where \( W \) and \( H \) are the width and height of the image \( I_{\text{new}} \), respectively, and \( i, j \) are pixel indices.

    \item \textbf{Decision Making:} 
    Check the computed DRIPS against the thresholds determined in the Training Phase, specifically, whether there holds:
    \begin{equation}
    L_{\text{lower}} \leq \text{DRIPS}(I_{\text{new}}) \leq L_{\text{upper}}
    \end{equation}
    In this case discard the data point as it's deemed not informative. Otherwise, retain the data point as it's considered important or informative for potential retraining or further analysis.
\end{enumerate}

By comparing the DRIPS of a new data point with the established thresholds, this phase effectively filters out routine or redundant data, focusing on capturing potentially informative or anomalous data points. The computational overhead on-device is minimal, primarily involving the computation of Grad-CAM heatmaps and DRIP scores, with a complexity of \(O(N)\). The more significant overhead occurs during the Training Phase, which includes model training and threshold determination. These intensive computations can be outsourced to more capable computational environments. To reimplement the code, please see the pseudocode.

\subsection{Pseudocode}
\begin{algorithm}
\caption{Selective Data Retention using DRIP algorithm}
\begin{algorithmic}
\Require
    \State TrainDataset: Dataset used for training the model
    \State TestDataset: Dataset used for evaluating the model
    \State Model: Neural network model
    \State DPW: Discard Percentage Window size in \%.
    \State NewDataPoint: New data point encountered in production

\Ensure 
    \State Decision: Whether to retain the NewDataPoint or not

    \State Train Model using TrainDataset
    \State Initialize empty list: DRIPS\_List
    \For{each Image in TestDataset}
        \State Compute Grad-CAM heatmap for Datapoint
        \State Compute DRIP Score for Datapoint:
        \State \[ \text{DRIP Score} = \frac{\sum_{i=1}^{W} \sum_{j=1}^{H} H(i,j)}{W \times H} \]
        \State Append DRIP Score to DRIP\_List
    \EndFor
    \State Create histograms of DRIPS\_List
    \State Determine peak of the histograms
    \State Calculate lower and upper thresholds for DPW data around the peaks of the histograms
    \State Compute Grad-CAM heatmap for NewDataPoint
    \State Compute DRIPS for NewDataPoint of the corresponding class
    \If{DRIPS is between the lower and upper thresholds}
        \State Decision = "Discard"
    \Else
        \State Decision = "Retain"
    \EndIf
    \State \Return Decision
\end{algorithmic}
\end{algorithm}

\section{Experimental Setup}

\subsection{Objective}


The primary aim of our experiment is to validate the effectiveness of the DRIP algorithm in enhancing model performance and ensuring efficiency in data storage.

\subsubsection{Datasplit}
\begin{itemize}
    \item \textbf{Training Dataset 40\%:} A labeled dataset utilized for the initial training of the model.
    \item \textbf{Validation Dataset 20\%:} A separate labeled set to test the model and generate DRIP Scores.
    \item \textbf{Production Dataset 40\%:} Simulated or real-world unlabeled data points that the model will encounter in a production-like scenario.
\end{itemize}

\subsubsection{Experimental Setup}
\paragraph{Baseline Model}
Train a neural network model using the entire training dataset (dataset 1) and evaluate its performance on a separate test dataset (dataset 2) to establish a baseline accuracy.

\paragraph{DRIP Model}
Train the model using the training dataset (dataset 1). Apply the proposed algorithm on the training dataset to determine the DRIPS thresholds. Simulate a production environment and apply the algorithm to the production dataset (dataset 3) to decide which data points to retain. Retrain the model using the retained data points and evaluate its performance on the test dataset (dataset 2).

\paragraph{All-Data Model}
Train a neural network model using the entire training dataset (dataset 1) + production dataset (dataset 3) and evaluate its performance on a separate test dataset (dataset 2) to establish an accuracy to compare our method.

\subsubsection{Evaluation Metrics}
\begin{itemize}
    \item \textbf{Model Performance:} Metrics such as accuracy, F1-score, etc., on the test dataset.
    \item \textbf{Data Retention Rate:} Percentage of data points retained from the production dataset.
    \item \textbf{Computational Efficiency:} Time taken for each data point's data retention decision.
    \item \textbf{Storage Savings:} Amount of storage saved due to selective data retention.
\end{itemize}

\subsubsection{Procedure}
\begin{figure}[!hbt] 
	\centering
	\includegraphics[width=0.48\textwidth]{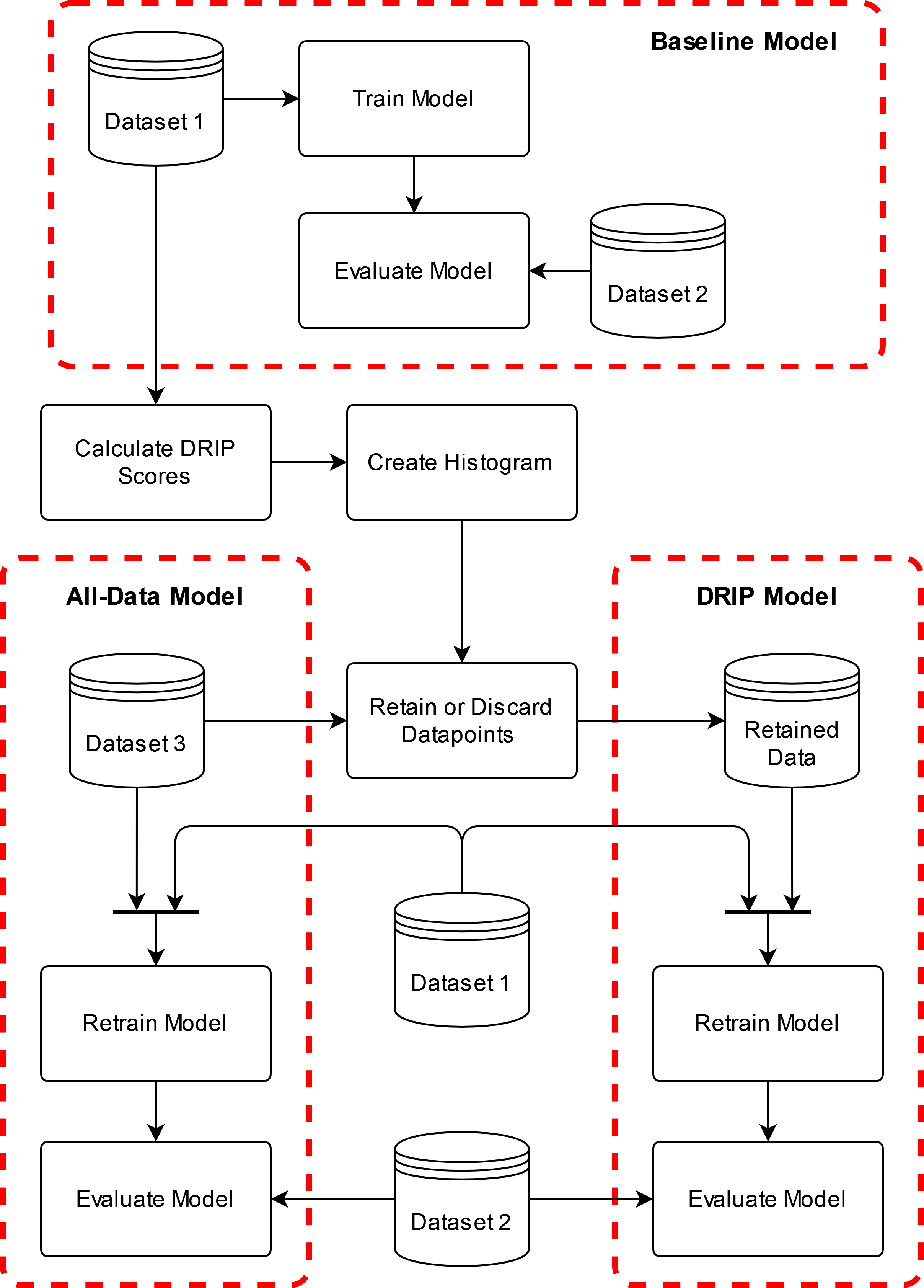} 
    \caption{Schematic representation of the experimental process detailing the computation of the three key metrics: Baseline Model Accuracy, All-Data Model Accuracy, and DRIP Model Accuracy}
    
    \label{fig:Exp_process}
    \end{figure}

In order to generate meaningful results, each experiment was carried out 20 times for each data set. In the results table \ref{tab:results_summary}, we show the average and the standard deviation of the results. Before each run, all data from the use case was randomly assigned to the 3 datasets. Afterward, the following 4 steps are carried out (this process is shown in Fig.~\ref{fig:Exp_process}):
\begin{enumerate}
    \item \textbf{Train the Baseline Model:} Train the model using the entire training dataset (dataset 1) and evaluate its performance on the test dataset (dataset 2).
    \item \textbf{Apply the DRIP algorithm:} Determine the DRIP Score thresholds using dataset 1 and decide which data points to retain from the production dataset (dataset 3).
    \item \textbf{Retrain and Evaluate:} Retrain the model using the retained data points and evaluate its performance on the test dataset (dataset 2).
    \item \textbf{Comparison:} Analyse the performance of the retrained model against the baseline and the accuracy with retraining with all data, considering data retention rate, computational efficiency, and storage efficiency.
\end{enumerate}

\subsection{Neural Network Architectures Used}

To evaluate the DRIP algorithm across different modalities and datasets, we utilized two distinct neural network architectures tailored to the nature of each dataset: an image-based model for visual datasets (MNIST, CIFAR-10, and Plant Disease) and a one-dimensional convolutional model for audio data (Speech Commands).

\subsubsection{Image-Based Datasets (MNIST, CIFAR-10, Plant Disease)}
For the image-based datasets, we employed \textbf{EfficientNet-B0}\cite{Tan.2019}, a lightweight yet powerful convolutional neural network that balances accuracy and computational efficiency, making it suitable for TinyML applications. The model was initialized with pretrained ImageNet weights and fine-tuned for each dataset's specific classification task.

To adapt EfficientNet-B0 for our purposes:
\begin{itemize}
    \item The final classification layer was replaced with a new fully connected layer matching the number of classes of the respective dataset.
    \item To reduce computational cost and improve generalization, we froze all layers of the EfficientNet-B0 feature extractor except for the last seven submodules (PyTorch). These include the final high-level convolutional blocks, the `ConvHead`, batch normalization, and the classifier head. This partial fine-tuning strategy retains low-level pretrained features while allowing the model to adapt to the target domain.

\end{itemize}

\subsubsection{Audio Dataset (Speech Commands)}
For the \textbf{Speech Commands} dataset, which consists of 1D audio data, we implemented a custom \textbf{1D Convolutional Neural Network (1D-CNN)} architecture. The model processes raw audio signals using a series of convolutional and pooling layers followed by fully connected layers for classification.

The architecture consists of:
\begin{itemize}
    \item Four 1D convolutional layers with batch normalization, ReLU activations, and max pooling operations.
    \item A fully connected head with a dropout layer and two linear layers, the final one projecting to the number of target classes.
\end{itemize}

These architectures were selected to reflect practical TinyML deployment scenarios while maintaining strong classification performance across a variety of input types.

\section{Results}
Our evaluation of the DRIP algorithm across four benchmark datasets demonstrates its efficacy in selective data retention, optimizing storage, and maintaining or enhancing model accuracy. This section presents our findings, focusing on storage savings, the impact of varying DPW size, and the algorithm's robustness to noise.
\begin{table*}[h]
\centering
\caption{Summary of results for MNIST, CIFAR-10, Plant Disease (PD) and Speech Commands (SC) datasets. The mean and standard deviation of the 20 runs is calculated. }
\begin{tabular}{|c|c|c|c|c|c|}
\hline
\textbf{Dataset} & \textbf{Baseline Acc.} & \textbf{All-Data Acc.} & \textbf{DRIP Acc.} & \textbf{Random Acc.} & \textbf{Storage Savings} \\
\hline
MNIST & \parbox[c][25pt]{2cm}{97.8\% \\ ($\pm 0.2\%$)} & \parbox[c][25pt]{2cm}{98.9\% \\ ($\pm 0.1\%$)} & \parbox[c][25pt]{2cm}{98.9\% \\ ($\pm 0.1\%$)} & \parbox[c][25pt]{2cm}{98.2\% \\ ($\pm 0.5\%$)} & \parbox[c][25pt]{2cm}{39\% \\ ($\pm 1\%$)} \\
\hline
CIFAR-10 & \parbox[c][25pt]{2cm}{87.5\% \\ ($\pm 0.3\%$)} & \parbox[c][25pt]{2cm}{89.1\% \\ ($\pm 0.2\%$)} & \parbox[c][25pt]{2cm}{89.2\% \\ ($\pm 0.2\%$)} & \parbox[c][25pt]{2cm}{88.8\% \\ ($\pm 0.6\%$)} & \parbox[c][25pt]{2cm}{23\% \\ ($\pm 2\%$)} \\
\hline
PD & \parbox[c][25pt]{2cm}{71.0\% \\ ($\pm 0.2\%$)} & \parbox[c][25pt]{2cm}{77.5\% \\ ($\pm 0.2\%$)} & \parbox[c][25pt]{2cm}{77.4\% \\ ($\pm 0.2\%$)} & \parbox[c][25pt]{2cm}{73.7\% \\ ($\pm 0.5\%$)} & \parbox[c][25pt]{2cm}{35\% \\ ($\pm 1\%$)} \\
\hline
SC & \parbox[c][25pt]{2cm}{76.6\% \\ ($\pm 0.4\%$)} & \parbox[c][25pt]{2cm}{85.6\% \\ ($\pm 0.5\%$)} & \parbox[c][25pt]{2cm}{85.7\% \\ ($\pm 0.4\%$)} & \parbox[c][25pt]{2cm}{80.1\% \\ ($\pm 0.7\%$)} & \parbox[c][25pt]{2cm}{29\% \\ ($\pm 3\%$)} \\
\hline
\end{tabular}
\label{tab:results_summary}
\end{table*}
\subsection{DRIP's Impact on Storage Efficiency}
The DRIP algorithm significantly improved storage efficiency across all tested datasets by selectively retaining only the most informative data points, as shown in Table \ref{tab:results_summary}. This approach not only preserved but sometimes enhanced model accuracy compared to using all available data.

Storage savings, on the other hand, represents the percentage reduction in the amount of data stored during the model's training or retraining process on an Edge Device. By selectively retaining only the most informative data points, the DRIP algorithm reduces the total amount of data that needs to be stored on the device, thereby saving storage space. This is particularly important for on-device training in resource-constrained environments, where memory is limited.
Here is DRIP's impact on each dataset:

\textbf{CIFAR-10:} Achieved 89.2\% mean accuracy, slightly higher than the all-data model's 89.1\%, with a 23\% mean reduction in storage.

\textbf{MNIST:} Matched the all-data model's 98.9\%  mean accuracy with a 39\% mean storage saving, demonstrating effectiveness in high-accuracy datasets.

\textbf{Speech Commands:} Slightly outperformed the all-data model (85.7\% vs. 85.6\%) with a 29\% mean storage reduction, indicating adaptability to audio data.

\textbf{Plant Disease:} Closely matched the all-data model's 77.4\% mean accuracy with a 35\% mean reduction in storage, highlighting potential in storage-constrained applications.

These results illustrate DRIP's capability to maintain or enhance model accuracy across diverse datasets while significantly reducing storage requirements, making it valuable for on-device applications with limited storage and computational resources.

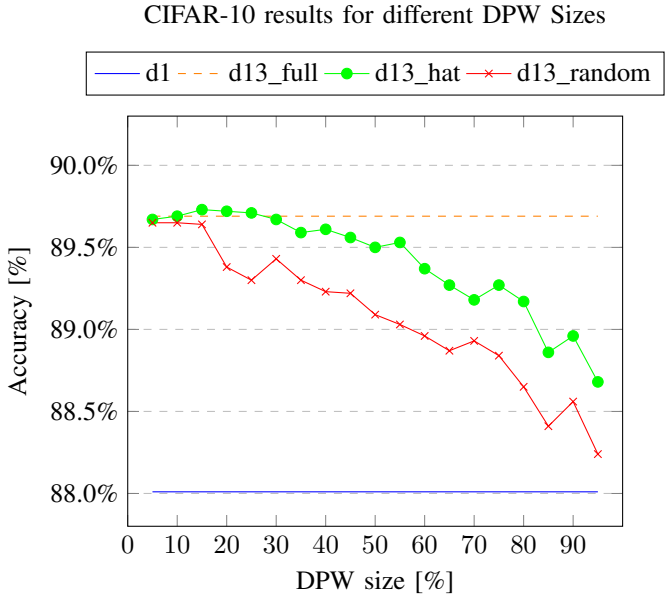
\begin{figure}[h]
\centering
\begin{tikzpicture}
\begin{axis}[
    title={CIFAR-10 results for different DPW Sizes},
    title style={yshift=0.9cm}, 
    xlabel={DPW size [\%]},
    ylabel={Accuracy [\%]},
    xmin=0, xmax=100,
    ymin=0.873, ymax=0.898,
    xtick={0,10,20,30,40,50,60,70,80,90},
    yticklabels={87.3\%,87.5\%,88.0\%,88.5\%,89.0\%,89.5\%,90.0\%},
    legend columns=-1,
    legend style={at={(0.5,1.05)},anchor=south},
    legend entries={d1,d13\_full,d13\_hat,d13\_random},
    ymajorgrids=true,
    grid style=dashed,
]

\addplot[
    color=blue,
    mark=none,
    error bars/.cd,
    y dir=both, y explicit,
]
coordinates {
    (5,0.8751)
    (10,0.8751)
    (15,0.8751)
    (20,0.8751)
    (25,0.8751)
    (30,0.8751)
    (35,0.8751)
    (40,0.8751)
    (45,0.8751)
    (50,0.8751)
    (55,0.8751)
    (60,0.8751)
    (65,0.8751)
    (70,0.8751)
    (75,0.8751)
    (80,0.8751)
    (85,0.8751)
    (90,0.8751)
    (95,0.8751)
};
\addlegendentry{d1}

\addplot[
    color=orange,
    dashed,
    mark=none,
    error bars/.cd,
    y dir=both, y explicit,
]
coordinates {
    (5,0.8919)
    (10,0.8919)
    (15,0.8919)
    (20,0.8919)
    (25,0.8919)
    (30,0.8919)
    (35,0.8919)
    (40,0.8919)
    (45,0.8919)
    (50,0.8919)
    (55,0.8919)
    (60,0.8919)
    (65,0.8919)
    (70,0.8919)
    (75,0.8919)
    (80,0.8919)
    (85,0.8919)
    (90,0.8919)
    (95,0.8919)
};
\addlegendentry{d13\_full}

\addplot[
    color=green,
    mark=*,
    error bars/.cd,
    y dir=both, y explicit,
]
coordinates {
    (5,0.8917)
    (10,0.8919)
    (15,0.8923)
    (20,0.8922)
    (25,0.8921)
    (30,0.8917)
    (35,0.8909)
    (40,0.8911)
    (45,0.8906)
    (50,0.89)
    (55,0.8903)
    (60,0.8887)
    (65,0.8877)
    (70,0.8868)
    (75,0.8877)
    (80,0.8867)
    (85,0.8836)
    (90,0.8846)
    (95,0.8818)
};
\addlegendentry{d13\_hat}

\addplot[
    color=red,
    mark=x,
    error bars/.cd,
    y dir=both, y explicit,
]
coordinates {
    (5,0.8915)
    (10,0.8915)
    (15,0.8914)
    (20,0.8888)
    (25,0.8880)
    (30,0.8893)
    (35,0.8880)
    (40,0.8873)
    (45,0.8872)
    (50,0.8859)
    (55,0.8853)
    (60,0.8846)
    (65,0.8837)
    (70,0.8843)
    (75,0.8834)
    (80,0.8815)
    (85,0.8791)
    (90,0.8806)
    (95,0.8774)
};
\addlegendentry{d13\_random}
\end{axis}
\end{tikzpicture}
\caption{Analysis of CIFAR-10 Model Accuracy Across Various DPW sizes: This graph compares the accuracy of different configurations (d1, d13\_full, d13\_hat, d13\_random) as the DPW size changes, highlighting the algorithm's sensitivity to parameter adjustments and its impact on model accuracy.}

\label{fig:cifarthp}
\end{figure}

\subsection{Investigation of Different DPW Sizes}

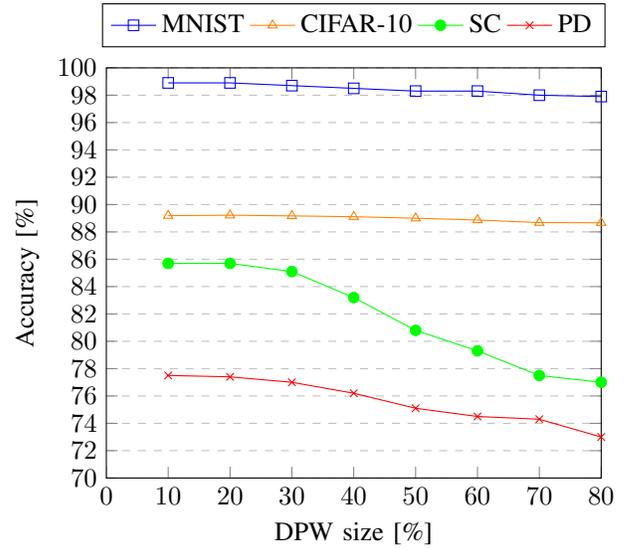
\begin{figure}[h]
\centering
\begin{tikzpicture}
\begin{axis}[
    title={Dataset Accuracy vs. DPW size},
    title style={yshift=0.9cm}, 
    xlabel={DPW size [\%]},
    ylabel={Accuracy [\%]},
    xmin=0, xmax=80,
    ymin=70, ymax=100,
    xtick={0,10,20,30,40,50,60,70,80},
    ytick={70,72,74,76,78,80,82,84,86,88,90,92,94,96,98,100},
    legend columns=-1,
    legend style={at={(0.5,1.05)},anchor=south},
    legend entries={MNIST,CIFAR-10,SC,PD},
    ymajorgrids=true,
    grid style=dashed,
]

\addplot[
    color=blue,
    mark=square,
    error bars/.cd,
    y dir=both, y explicit,
]
coordinates {
    (10,98.9)
    (20,98.9)
    (30,98.7)
    (40,98.5)
    (50,98.3)
    (60,98.3)
    (70,98.0)
    (80,97.9)
};
\addlegendentry{MNIST}

\addplot[
    color=orange,
    mark=triangle,
    error bars/.cd,
    y dir=both, y explicit,
]
coordinates {
    (10,89.19)
    (20,89.22)
    (30,89.17)
    (40,89.11)
    (50,89)
    (60,88.87)
    (70,88.68)
    (80,88.67)
};
\addlegendentry{CIFAR-10}

\addplot[
    color=green,
    mark=*,
    error bars/.cd,
    y dir=both, y explicit,
]
coordinates {
    (10,85.7)
    (20,85.7)
    (30,85.1)
    (40,83.2)
    (50,80.8)
    (60,79.3)
    (70,77.5)
    (80,77)
};
\addlegendentry{SC}

\addplot[
    color=red,
    mark=x,
    error bars/.cd,
    y dir=both, y explicit,
]
coordinates {
    (10,77.5)
    (20,77.4)
    (30,77.0)
    (40,76.2)
    (50,75.1)
    (60,74.5)
    (70,74.3)
    (80,73)
};
\addlegendentry{PD}
\end{axis}
\end{tikzpicture}
\caption{Model Accuracy Across Datasets with Varying DPW sizes: Demonstrates the effect of different DPW sizes on the accuracy of the datasets.}

\label{fig:THP}
\end{figure}

Our analysis extends to exploring different Discard Percentage Window (DPW) sizes, examining their impact on model accuracy. The CIFAR-10 dataset, for example, showed in Figure \ref{fig:cifarthp} a slight improvement in accuracy with DRIP over the all-data model, highlighting the algorithm's adeptness at handling varying levels of data retention. Figure \ref{fig:THP} illustrates the model accuracy across different DPW sizes, showcasing DRIP's consistent performance even as the bandwidth adjustments are made.

\subsection{Robustness to Noise}
To evaluate the robustness of DRIP (DRop unImportant data Points) against noisy data, we conducted experiments on the CIFAR-10 dataset by introducing noise and mislabeling. We incrementally added noise in steps of 10\% and retrained the model twice at each noise level: once with the unfiltered dataset and once with the dataset filtered by DRIP.

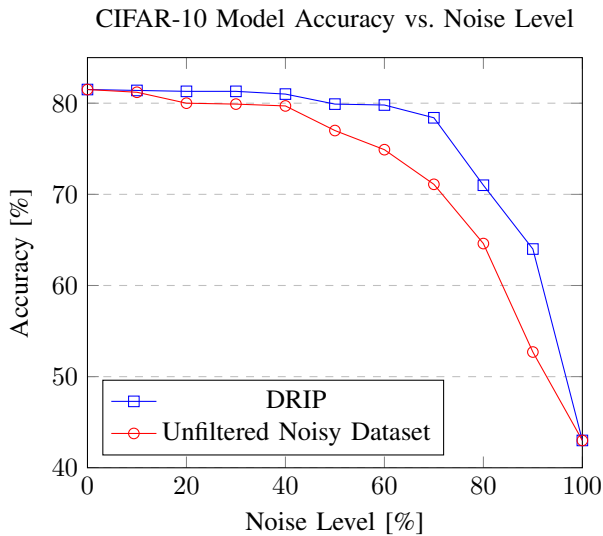
\begin{figure}[h]
\centering
\begin{tikzpicture}
\begin{axis}[
    title={CIFAR-10 Model Accuracy vs. Noise Level},
    xlabel={Noise Level [\%]},
    ylabel={Accuracy [\%]},
    xmin=0, xmax=100,
    ymin=40, ymax=85,
    legend pos=south west,
    ymajorgrids=true,
    grid style=dashed,
]

\addplot[
    color=blue,
    mark=square,
    ]
    coordinates {
    (0,81.5)(10,81.4)(20,81.3)(30,81.3)(40,81)(50,79.9)(60,79.8)(70,78.4)(80,71)(90,64)(100,43)
    };
    \addlegendentry{DRIP}

\addplot[
    color=red,
    mark=o,
    ]
    coordinates {
    (0,81.5)(10,81.2)(20,80)(30,79.9)(40,79.7)(50,77)(60,74.9)(70,71.1)(80,64.6)(90,52.7)(100,43)
    };
    \addlegendentry{Unfiltered Noisy Dataset}


\end{axis}
\end{tikzpicture}
\caption{Model Accuracy Comparison under Increasing Noise Levels - Demonstrating DRIP's superior performance in maintaining accuracy against unfiltered noisy data in CIFAR-10.}
\label{fig:noise_vs_accuracy}

\end{figure}

The results, shown in Figure \ref{fig:noise_vs_accuracy}, illustrate that as noise increases, the accuracy of the model trained on the unfiltered dataset declines more sharply compared to the model trained with data filtered by DRIP. These findings demonstrate DRIP's effectiveness in handling noisy data by selectively retaining more reliable and informative data points, thus enhancing overall model performance and resilience.

\section{Discussion}

The evaluation of the DRIP algorithm across diverse datasets highlights its effectiveness in selective data retention and storage optimization, while maintaining or even enhancing model accuracy. This discussion elaborates on the key findings and provides a deeper analysis of the implications, supported by specific evidence from the results.

\subsection{Model Performance and Efficiency}

The DRIP algorithm demonstrated strong performance across all tested datasets, including CIFAR-10, MNIST, Speech Commands, and Plant Disease. In terms of accuracy, the algorithm consistently matched or exceeded models trained on the entire dataset. For instance, in CIFAR-10, DRIP achieved a higher accuracy (89.2\%) than the all-data model (89.1\%), while for MNIST, it achieved the same 98.9\% accuracy as the all-data model. This suggests that selectively retaining data based on the DRIP Score does not compromise, and may even improve, model performance.

The slight improvement in accuracy observed in CIFAR-10 and Speech Commands suggests that DRIP’s selective retention of the most informative data points may improve the model’s ability to generalize. By training on a refined subset of data that contributes meaningfully to learning, the algorithm prevents overfitting and ensures the model is exposed to data points that drive learning outcomes. This aligns with the general observation that data quality often trumps quantity in machine learning.

\subsection{Efficiency in Data Retention and Storage Savings}

One of the key benefits of the DRIP algorithm is its ability to achieve significant storage savings without compromising performance. Across the four datasets, DRIP selectively retained between 61\% and 77\% of the original data, resulting in storage savings ranging from 23\% to 39\%. For instance, the MNIST dataset showed a 39\% reduction in stored data while maintaining the same accuracy as the all-data model. This level of efficiency is especially valuable in resource-constrained environments such as edge computing or IoT devices, where storage is at a premium.

The storage savings do not come at the cost of learning efficacy. In fact, by filtering out redundant or less informative data points, DRIP ensures that the model is retrained on high-quality data, which contributes directly to model performance. This is particularly advantageous for on-device applications, where not only storage but also computational power is limited.

\subsection{Adaptability and Streaming Decision Making}

A notable strength of DRIP is its adaptability to a variety of data types. Whether applied to image datasets like CIFAR-10 and MNIST, or audio data like Speech Commands, DRIP consistently performed well. This versatility makes the algorithm a promising solution for different domains and data modalities, from image classification to speech recognition.

Moreover, DRIP’s ability to make on-device decisions about data retention is critical for applications requiring immediate processing, such as autonomous systems and mobile devices. The Grad-CAM-based scoring system allows the algorithm to assess the importance of incoming data points and determine whether they should be retained or discarded in real time. This dynamic data management process supports continuous learning and adaptation in scenarios where fast, on-device retraining is necessary.

\subsection{General Observations}

Several general observations from the experimental results highlight the broader advantages of the DRIP algorithm:

    Consistent Performance Across Datasets: DRIP's performance across diverse datasets underscores the robustness of the approach, making it applicable to a wide range of machine learning tasks and environments.
    Quality over Quantity: The consistent accuracy of models trained with selectively retained data points illustrates the importance of focusing on high-quality, informative data rather than large volumes of data.
    On-device Efficiency: The ability to make data retention decisions with streaming data offers practical advantages for on-device machine learning, where both storage and response time are critical.

\subsection{Limitations and Future Directions}

While DRIP offers substantial advantages, there are limitations to consider. The initial Training Phase, where thresholds for DRIP scores are determined, can be computationally intensive, particularly when working with large datasets. This phase could benefit from optimization, such as incorporating parallel processing or offloading it to cloud environments.

Additionally, the algorithm’s effectiveness depends on the quality of the initial dataset. If the training data used to calculate DRIP scores is not representative of the overall data distribution, the algorithm may retain uninformative or biased data points, leading to suboptimal performance. Future research could explore methods to dynamically adjust the DRIP score thresholds or integrate other visualization techniques to further enhance data selection accuracy.

Furthermore, real-world deployments of DRIP on edge devices would provide valuable insights into its practical applications, particularly in dynamic and evolving environments where data characteristics change over time.

\subsection{Broader Impacts and Ethical Considerations}

DRIP’s efficient data retention approach has the potential to significantly reduce storage and computational costs, making advanced machine learning technologies more accessible and sustainable, particularly in resource-constrained environments. However, the selective retention process may introduce biases if the initial training data is not representative. This could result in models that perform well for certain data subgroups but poorly for others. Ensuring that the training datasets are diverse and representative is crucial to mitigating this risk. In addition, regular monitoring of algorithmic performance across various demographics and data sources is essential to ensure fairness and avoid unintended biases in decision-making.

\section{Conclusion}

We introduced and evaluated the DRIP algorithm for streaming selective data retention, addressing challenges in on-device machine learning, particularly in TinyML with limited resources.

Our experimental results show that DRIP achieves near-identical performance to models trained on the entire dataset while ensuring significant storage savings. This demonstrates DRIP's ability to retain only the most informative data points, optimizing storage without compromising performance.

Implementing DRIP enables devices to operate more efficiently, reducing data transmissions and extending operational lifespans. The reduced data storage requirements also lead to cost savings in storage and data management.

In conclusion, the DRIP algorithm represents a significant advancement in efficient on-device machine learning, balancing performance and efficiency for next-generation TinyML applications.

\bibliographystyle{IEEEtran}
\bibliography{refs} 
\end{document}